\documentclass[conference]{IEEEtran}
\IEEEoverridecommandlockouts

\usepackage{cite}
\usepackage{amsmath,amssymb,amsfonts}
\usepackage{algorithmic}
\usepackage{graphicx}
\usepackage{textcomp}
\usepackage{xcolor}
\def\BibTeX{{\rm B\kern-.05em{\sc i\kern-.025em b}\kern-.08em
    T\kern-.1667em\lower.7ex\hbox{E}\kern-.125emX}}
\begin{document}

\title{A CUBS-Compatible Ultrasound Morphology and Uncertainty-Aware Baseline for Carotid Intima–Media Segmentation and Preliminary Risk Prediction\\

}

\author{

\IEEEauthorblockN{Aueaphum Aueawatthanaphisut}
\IEEEauthorblockA{
\textit{School of Information, Computer, and Communication Technology} \\
\textit{Sirindhorn International Institute of Technology, Thammasat University} \\
Pathum Thani, Thailand \\
aueawatth.aue@gmail.com
}
}

\maketitle

\begin{abstract}
Carotid atherosclerosis is a major contributor to ischemic stroke and transient ischemic attack; however, conventional ultrasound-based assessment remains largely dependent on morphology- and velocity-centered indicators such as intima--media thickness, plaque appearance, stenosis degree, and peak systolic velocity. These indicators may be insufficient to fully characterize patient-specific vascular risk because atherosclerosis progression is strongly influenced by local hemodynamic disturbance, wall shear stress distribution, and uncertainty in image-derived interpretation. In this study, an uncertainty-aware physics-guided multiscale learning framework, termed \textit{AtheroFlow-XNet}, is proposed for patient-specific carotid atherosclerosis risk stratification. The framework is designed to integrate ultrasound-derived vascular morphology, clinical descriptors, uncertainty-aware prediction, and an extensible hemodynamic learning formulation that can incorporate Doppler-informed flow profiles and computational fluid dynamics-derived wall shear biomarkers when available. Using the Carotid Ultrasound Boundary Study (CUBS) dataset, the proposed model was implemented as a morphology-driven baseline for carotid intima--media segmentation and risk prediction. Manual lumen--intima and media--adventitia boundary annotations were converted into dense intima--media masks to supervise the segmentation branch, while clinical variables were incorporated into the risk-prediction branch. The model achieved a Dice coefficient of 0.7930 for LI--MA mask segmentation, a segmentation loss of 0.2359, and an area under the receiver operating characteristic curve of 0.6910 for risk prediction. Qualitative analysis showed that predicted masks were generally aligned with manual annotations, and Monte Carlo dropout-based uncertainty maps highlighted ambiguous wall-boundary regions. These findings suggest that ultrasound-derived carotid morphology can be effectively learned for automated wall analysis, while uncertainty-aware inference can provide additional interpretability for clinically difficult cases. The present study establishes a reproducible CUBS-compatible foundation for \textit{AtheroFlow-XNet}; future work will incorporate Doppler-derived boundary conditions, patient-specific vascular reconstruction, and CFD-based hemodynamic biomarkers to complete the proposed physics-guided risk stratification pipeline.
\end{abstract}


\section{Introduction}

Carotid atherosclerosis has been recognized as one of the major pathological contributors to ischemic stroke and transient ischemic attack. Although the degree of luminal stenosis has traditionally been used as the principal criterion for clinical decision-making, it has been increasingly reported that stenosis severity alone is insufficient to represent plaque vulnerability and patient-specific cerebrovascular risk [3,10,12]. In clinical practice, a considerable number of carotid plaques may remain asymptomatic despite severe narrowing, whereas non-severe stenotic lesions may still be associated with future neurological events. Therefore, additional physiological and biomechanical markers are required to characterize the local vascular environment beyond purely anatomical measurements [5,20].

Hemodynamic forces have been widely implicated in the initiation, progression, and destabilization of atherosclerotic plaques. Among these forces, wall shear stress (WSS), time-averaged wall shear stress (TAWSS), oscillatory shear index (OSI), and relative residence time (RRT) have been frequently investigated as mechanobiological indicators of disturbed flow and endothelial dysfunction [4,5,19,20]. Regions exposed to low and oscillatory WSS have been associated with plaque initiation and local wall thickening, whereas elevated WSS near stenotic throats has been associated with plaque rupture susceptibility and ulceration in advanced lesions [18,20]. These findings suggest that carotid plaque risk should be interpreted as a spatially heterogeneous and temporally dynamic process rather than as a single geometric index.

\begin{figure*}[h]
    \centering
    \includegraphics[width=\textwidth]{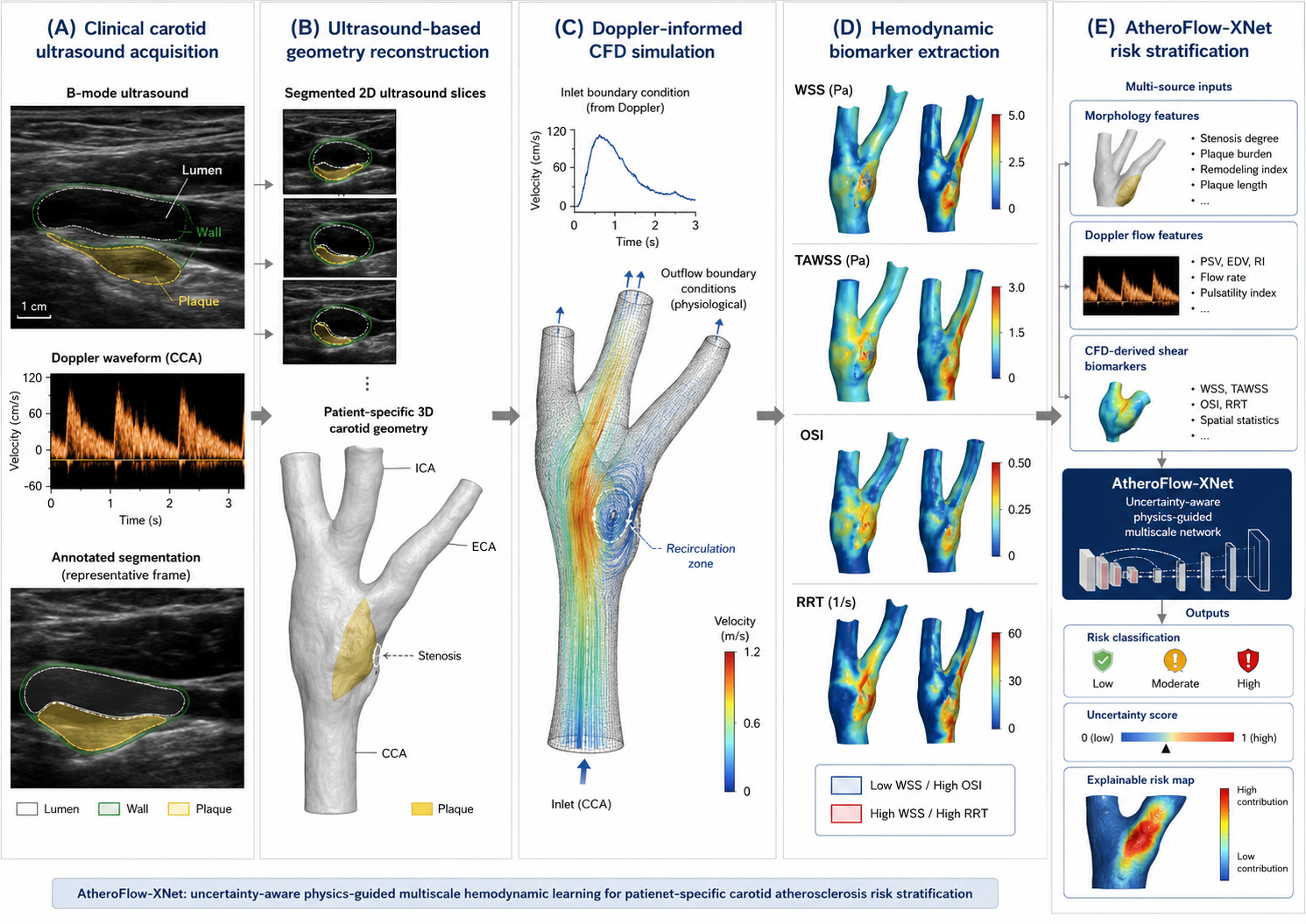}
    \caption{
    Overall workflow of the proposed \textit{AtheroFlow-XNet} framework for patient-specific carotid atherosclerosis risk stratification. 
    (A) B-mode carotid ultrasound images and Doppler-derived velocity waveforms are acquired to characterize vascular morphology and pulsatile flow dynamics. Lumen, vessel wall, and plaque boundaries are annotated to provide image-based anatomical information for downstream modeling. 
    (B) Segmented two-dimensional ultrasound slices are transformed into a patient-specific three-dimensional carotid bifurcation geometry, including the common carotid artery (CCA), internal carotid artery (ICA), external carotid artery (ECA), plaque region, and stenotic segment. 
    (C) Doppler-informed computational fluid dynamics (CFD) simulation is performed using patient-specific inlet flow profiles and physiologically consistent outlet boundary conditions to estimate local velocity fields, streamline patterns, and recirculation zones. 
    (D) Multiscale hemodynamic biomarkers, including wall shear stress (WSS), time-averaged wall shear stress (TAWSS), oscillatory shear index (OSI), and relative residence time (RRT), are extracted as surface-based risk maps to identify disturbed-flow regions, low-WSS plaque-prone zones, and high-WSS stenotic throat regions. 
    (E) Morphological features, Doppler-derived flow descriptors, and CFD-based shear biomarkers are integrated into the proposed uncertainty-aware physics-guided multiscale learning model to generate individualized low-, moderate-, or high-risk classification, uncertainty estimation, and explainable hemodynamic risk maps.
    }
    \label{fig:atheroflow_overview}
\end{figure*}

Computational fluid dynamics (CFD) has been increasingly employed to quantify patient-specific carotid hemodynamics because it enables the estimation of velocity fields, pressure distributions, WSS-related biomarkers, and recirculation patterns that are difficult to measure directly in vivo [4,6,7]. Patient-specific CFD analyses have demonstrated that adverse WSS and OSI distributions are spatially correlated with plaque burden, plaque composition, and changes in plaque thickness [4,19]. Moreover, stenosis severity, stenosis location, bifurcation angle, and plaque morphology have been shown to substantially alter local hemodynamic environments, thereby affecting both plaque formation and rupture-prone conditions [15,16,17,18]. However, despite its mechanistic value, CFD remains limited in routine clinical translation because its accuracy is strongly dependent on vascular reconstruction quality, boundary condition selection, mesh generation, computational cost, and post-processing standardization [6,7].

Ultrasound imaging remains one of the most accessible and clinically practical modalities for carotid assessment because of its non-invasive nature, low cost, portability, and real-time acquisition capability [2,3,9,10]. Conventional B-mode and Doppler ultrasound have been commonly used to estimate intima--media thickness, plaque morphology, stenosis degree, and peak systolic velocity [2,12]. More recently, ultrasound vector flow imaging and ultrafast Doppler techniques have enabled more detailed quantification of flow direction, turbulence-related behavior, and WSS-derived parameters in carotid arteries [9,10,11,13]. These ultrasound-based hemodynamic measurements have shown promising value for identifying carotid stenosis, cardiovascular risk in type 2 diabetes, intraplaque neovascularization, and vulnerable plaque characteristics [9,10,11,12]. Nevertheless, ultrasound-derived measurements are still affected by image quality, operator dependency, limited field of view, and difficulties in directly recovering three-dimensional patient-specific flow structures.

To overcome these limitations, ultrasound-based CFD frameworks have recently been proposed for reconstructing carotid geometries and generating personalized hemodynamic risk maps [3]. In such frameworks, tracked two-dimensional ultrasound images and automated segmentation can be used to reconstruct patient-specific vascular geometries, after which CFD simulations can be performed to quantify TAWSS, OSI, RRT, and helicity patterns [3]. This approach demonstrates that ultrasound-driven vascular reconstruction can serve as a practical bridge between clinical imaging and mechanistic hemodynamic analysis. However, most existing studies remain limited by small cohorts, simplified learning pipelines, deterministic outputs, and insufficient integration of multimodal information such as B-mode morphology, Doppler-derived flow profiles, CFD-based shear biomarkers, and uncertainty estimates [3,6,7].

Recent advances in machine learning and physics-informed modeling have provided new opportunities for accelerating cardiovascular flow analysis and improving patient-specific risk prediction. Physics-informed neural networks and hybrid CFD--learning frameworks have been introduced to preserve governing physical constraints while reducing the dependency on purely data-driven learning and extensive labeled datasets [21]. In cardiovascular simulations, such hybrid approaches have been used to infer boundary conditions, accelerate hemodynamic prediction, and support non-invasive diagnostic modeling under patient-specific anatomical variations [7,21]. However, the integration of ultrasound-derived vascular geometry, Doppler-informed flow boundary conditions, CFD-derived hemodynamic biomarkers, and uncertainty-aware risk learning has not yet been sufficiently established for carotid atherosclerosis stratification.

To address this limitation, the proposed \textit{AtheroFlow-XNet} framework was designed as a multimodal hemodynamic learning pipeline in which ultrasound-derived vascular morphology, Doppler-informed flow profiles, and CFD-based shear biomarkers are jointly modeled under physics-guided and uncertainty-aware constraints. The overall concept of the proposed framework is illustrated in Fig.~\ref{fig:atheroflow_overview}.

Therefore, in this study, an uncertainty-aware physics-guided multiscale hemodynamic learning framework, termed \textit{AtheroFlow-XNet}, is proposed for patient-specific atherosclerosis risk stratification. The proposed framework is designed to integrate ultrasound-reconstructed vascular geometry, Doppler-derived flow profiles, and CFD-based wall shear biomarkers within a unified predictive architecture. In contrast to conventional image-only or CFD-only approaches, the proposed model is intended to learn complementary morphological, flow-based, and mechanistic representations while preserving physiological consistency through physics-guided constraints. Uncertainty estimation is further incorporated to provide confidence-aware prediction, which is essential for clinical decision support in high-risk vascular assessment.

The main contributions of this study are summarized as follows. First, a multimodal patient-specific pipeline is formulated to combine B-mode ultrasound morphology, Doppler-derived flow information, and CFD-derived hemodynamic biomarkers for carotid risk analysis. Second, a physics-guided learning strategy is introduced to embed mechanistic vascular flow knowledge into the predictive model. Third, multiscale hemodynamic descriptors, including WSS, TAWSS, OSI, RRT, and disturbed-flow-related indicators, are incorporated to capture both local plaque-prone regions and global vascular risk patterns. Finally, an uncertainty-aware prediction mechanism is developed to support robust and interpretable patient-specific atherosclerosis risk stratification. Through this framework, carotid plaque assessment is expected to be shifted from stenosis-centered evaluation toward personalized, mechanism-informed, and clinically interpretable hemodynamic risk prediction.

\section{Related Work}

Carotid atherosclerosis risk assessment has traditionally been performed using anatomical and velocity-based indicators, including luminal stenosis severity, plaque morphology, carotid intima--media thickness (CIMT), and peak systolic velocity (PSV). Among these indicators, CIMT has been widely used as a surrogate marker of atherosclerosis and cardiovascular risk in B-mode ultrasound imaging. In the Carotid Ultrasound Boundary Study (CUBS), computerized CIMT measurement systems were evaluated using a large multicenter ultrasound dataset, and computerized measurements were shown to be comparable to expert manual segmentations for clinical outcome investigation [2]. Although such morphology-based measurements provide useful vascular information, they are insufficient to fully characterize plaque vulnerability and patient-specific hemodynamic risk.

\subsection{Ultrasound-Based Carotid Assessment}

Ultrasound imaging has remained one of the most practical modalities for carotid artery assessment because it is non-invasive, cost-effective, portable, and widely available in routine clinical practice. Conventional B-mode ultrasound has been used to assess plaque presence, vessel wall morphology, stenosis degree, and CIMT, whereas Doppler ultrasound has been used to estimate flow velocity and stenosis severity. However, conventional Doppler-based assessment is commonly limited by angle dependency, simplified velocity assumptions, and incomplete characterization of complex flow near bifurcations and stenotic plaques.

To overcome these limitations, ultrasound vector flow imaging (VFI) and vascular vector flow mapping have recently been investigated for quantitative carotid hemodynamic evaluation. In patients with type 2 diabetes mellitus, V-Flow imaging was used to quantify carotid wall shear stress (WSS), and reduced mean WSS was found to be associated with cardiovascular disease, suggesting that ultrasound-derived WSS may provide additional information for atherosclerosis evaluation [9]. In high-stroke-risk populations, turbulence index and WSS derived from high-frame-rate VFI were shown to improve the quantitative evaluation of internal carotid artery stenosis, with combined hemodynamic parameters achieving promising diagnostic performance [10]. Furthermore, V-flow-based WSS measurement was used to predict intraplaque neovascularization, and maximum WSS at the thickest plaque region was reported to be correlated with neovascularization enhancement grade [11]. These studies indicate that ultrasound-derived flow biomarkers can provide clinically meaningful information beyond morphology-only assessment.

Ultrafast ultrasound imaging has also been used for plaque vulnerability assessment by combining shear wave elastography and ultrafast Doppler-derived WSS. A multiparametric score based on plaque stiffness and flow analysis was reported to identify histologically vulnerable plaques with high diagnostic performance [12]. In addition, age-related changes in carotid WSS were quantified using vascular vector flow mapping in healthy adults, demonstrating the potential of ultrasound-based WSS measurement for establishing reference hemodynamic patterns [13]. Despite these advances, ultrasound-based WSS and flow measurements remain limited by field-of-view constraints, two-dimensional acquisition, operator dependency, and challenges in reconstructing full three-dimensional flow structures.

\subsection{Patient-Specific Computational Fluid Dynamics for Carotid Hemodynamics}

Computational fluid dynamics (CFD) has been increasingly used to estimate patient-specific vascular hemodynamics because velocity fields, pressure distributions, WSS, oscillatory shear index (OSI), time-averaged wall shear stress (TAWSS), relative residence time (RRT), and recirculation patterns can be computed with high spatial resolution. In patient-specific carotid artery models, low WSS and high OSI have been associated with plaque formation, plaque burden, and adverse vascular remodeling. LaDisa et al. demonstrated that regional carotid fluid dynamics could be quantified in a time-efficient patient-specific CFD framework, and that plaque thickness and plaque composition were spatially related to adverse WSS and OSI distributions [4]. Similarly, carotid geometry and WSS were shown to independently predict increased wall thickness in high-risk patients, supporting the clinical relevance of geometry--hemodynamic interactions in atherosclerosis progression [5].

The importance of realistic boundary conditions in carotid CFD has also been widely emphasized. The choice of inlet and outlet boundary conditions has been shown to influence simulated WSS, pressure, pressure gradient, and velocity fields. In a comparative assessment of boundary conditions for human carotid CFD, Windkessel, structured-tree, and fully developed outlet models were evaluated against clinical pressure and ultrasound Doppler velocity measurements, and physiologically informed boundary conditions were shown to improve simulation reliability [6]. An analytical method informed by clinical imaging data was later proposed for estimating carotid outlet boundary conditions, emphasizing that realistic distal vascular resistance and compliance are essential for accurate patient-specific flow simulation [7]. These findings suggest that Doppler-informed and physiologically consistent boundary conditions are necessary for clinically meaningful CFD-based carotid risk assessment.

Several CFD studies have further investigated the effects of stenosis severity, stenosis location, plaque morphology, bifurcation angle, and blood rheology on carotid hemodynamics. Non-Newtonian blood flow through the carotid bifurcation has been analyzed to identify sinus and bifurcation regions that are susceptible to atherosclerosis, where low mean WSS, secondary flow, and pressure-related effects were observed [14]. Plaque classification and bifurcation angle were also shown to modulate TAWSS, OSI, and endothelial cell activation potential, indicating that local geometry and plaque distribution strongly influence plaque vulnerability [15]. Numerical studies of carotid bifurcation stenosis demonstrated that stenosis degree alters pulsatility, pressure drop, velocity distribution, vorticity, and WSS [16]. Retrospective CFD analysis using vessel surface repairing further suggested that atherosclerosis-prone regions are characterized by low WSS, low velocity, and vortex formation before and after stenosis formation [17]. More recently, stenosis severity and location were shown to produce distinct WSS, OSI, and RRT patterns, indicating that hemodynamic risk is highly dependent on both anatomical and lesion-specific characteristics [18].

\subsection{Hemodynamic Biomarkers and Plaque Vulnerability}

The relationship between hemodynamic biomarkers and carotid plaque vulnerability has been extensively investigated. WSS has been considered a key mechanobiological stimulus because it represents the frictional force exerted by flowing blood on endothelial cells. Low and oscillatory WSS have been associated with endothelial dysfunction, plaque initiation, and local wall thickening, whereas high WSS in stenotic throats has been associated with plaque rupture, ulceration, and advanced plaque destabilization. In advanced human carotid atherosclerosis, TAWSS and OSI were correlated with histologically determined plaque composition, and larger necrotic core and macrophage areas were observed in regions exposed to high TAWSS or low OSI [19]. These findings suggest that the role of WSS may vary across plaque stages, with low WSS being involved in plaque initiation and high WSS being involved in advanced plaque destabilization.

Time-varying WSS has also been associated with plaque ulceration. In the Plaque at Risk Study, carotid plaques that developed ulcerations were compared with silent plaques, and high WSS and low RRT regions were found to have significantly higher odds of presenting ulceration even after adjustment for wall thickness [20]. This indicates that plaque vulnerability cannot be adequately represented by static stenosis degree alone. Instead, temporally varying shear-related biomarkers, such as WSS, OSI, and RRT, should be incorporated into patient-specific risk stratification. Collectively, these studies support the use of multiscale hemodynamic descriptors for identifying plaque-prone regions, disturbed-flow zones, and rupture-susceptible areas.

\subsection{Ultrasound-Driven CFD and Multimodal Hemodynamic Modeling}

Although CFD has provided detailed mechanistic insight into carotid hemodynamics, its clinical translation has been limited by the need for high-quality three-dimensional vascular geometry, appropriate boundary conditions, and substantial computational resources. To improve clinical applicability, ultrasound-driven CFD frameworks have recently been proposed. In an ultrasound-based CFD analysis of carotid artery hemodynamics, tracked two-dimensional ultrasound and automated segmentation were used to reconstruct carotid geometries in healthy and stenosed conditions, after which TAWSS, OSI, RRT, and helicity were quantified as risk maps [3]. This work demonstrated the feasibility of ultrasound-based patient-specific hemodynamic assessment and showed that stenosed carotid geometries exhibited extensive disturbed flow and altered helicity patterns compared with healthy arteries [3].

Despite this progress, existing ultrasound-driven CFD approaches have generally remained deterministic and simulation-centered. The integration of B-mode morphology, Doppler-derived flow profiles, CFD-derived shear biomarkers, and machine learning-based risk prediction has not yet been sufficiently established. Moreover, uncertainty estimation and physics-guided constraints have rarely been incorporated into carotid ultrasound--CFD pipelines. As a result, there remains a methodological gap between image-based anatomical assessment, physics-based hemodynamic simulation, and clinically interpretable risk prediction.

\subsection{Physics-Guided Learning and Hybrid CFD--AI Models}

Recent advances in physics-informed neural networks (PINNs), hybrid CFD--AI modeling, and physics-guided learning have introduced new opportunities for cardiovascular flow prediction. These approaches are designed to incorporate physical laws, boundary conditions, and conservation principles into machine learning models, thereby reducing reliance on purely data-driven training. A hybrid CFD--PINN--FSI framework was recently proposed for coronary artery simulations, where physics-informed modeling was used to estimate outlet flow conditions and improve patient-specific hemodynamic prediction under complex vascular geometries [21]. Although this work focused on coronary artery trees, its methodological principles are highly relevant to carotid hemodynamic modeling because outlet boundary estimation, flow conservation, and computational efficiency are also critical challenges in carotid CFD.

However, physics-guided AI has not yet been fully exploited for ultrasound-based carotid atherosclerosis risk stratification. Existing AI models for vascular assessment are often image-only or feature-only models, while CFD models are often computationally expensive and not directly optimized for prediction. Therefore, a unified framework that integrates ultrasound morphology, Doppler-informed flow, CFD-derived hemodynamic biomarkers, physics-guided constraints, and uncertainty-aware risk prediction is required.

\subsection{Research Gap}

Based on the above literature, four major limitations can be identified. First, conventional ultrasound-based assessment remains primarily morphology- and velocity-centered, and therefore may not fully capture three-dimensional disturbed-flow mechanisms. Second, CFD-based carotid analysis provides mechanistic biomarkers but is often limited by computational cost, boundary condition uncertainty, and lack of direct predictive learning. Third, ultrasound-driven CFD frameworks have demonstrated feasibility, but multimodal learning from morphology, Doppler flow, and CFD shear maps remains insufficiently developed. Fourth, uncertainty-aware and physics-guided AI models have rarely been applied to patient-specific carotid atherosclerosis risk stratification.

To address these limitations, the proposed \textit{AtheroFlow-XNet} framework is developed as an uncertainty-aware physics-guided multiscale hemodynamic learning model. In this framework, ultrasound-reconstructed vascular geometry, Doppler-derived flow profiles, and CFD-based wall shear biomarkers are jointly integrated to provide patient-specific risk classification, uncertainty estimation, and explainable hemodynamic risk mapping.

\section{Methodology}
\label{sec:methodology}

In this study, an uncertainty-aware physics-guided multiscale hemodynamic learning framework, termed \textit{AtheroFlow-XNet}, was formulated for patient-specific carotid atherosclerosis risk stratification. The proposed methodology was designed to extend conventional carotid ultrasound analysis from morphology-centered measurement toward a mechanism-informed predictive framework in which image-derived vascular structure, boundary-aware flow representation, and hemodynamic risk learning are jointly modeled. The overall training implementation in this work was based on the Carotid Ultrasound Boundary Study (CUBS) dataset, which provides B-mode common carotid artery ultrasound images, calibration factors, manual lumen--intima (LI) and media--adventitia (MA) boundary annotations, computerized segmentations, clinical variables, and long-term follow-up cardiovascular event information [2]. Because CUBS does not provide Doppler waveforms or CFD-derived WSS fields, the experimental implementation was focused on the ultrasound morphology, boundary segmentation, CIMT-derived geometric representation, clinical-risk prediction, and uncertainty-aware modules. The Doppler-informed CFD branch was mathematically specified as an extensible component to be activated when patient-specific Doppler and CFD data are available.

\subsection{Dataset Description and Problem Formulation}
\label{subsec:dataset_problem}

Let $\mathcal{D}=\{(\mathbf{I}_i,\mathbf{B}^{LI}_i,\mathbf{B}^{MA}_i,\mathbf{c}_i,y_i)\}_{i=1}^{N}$ denote the training dataset, where $\mathbf{I}_i\in\mathbb{R}^{H\times W}$ is a B-mode carotid ultrasound image, $\mathbf{B}^{LI}_i$ and $\mathbf{B}^{MA}_i$ are the manual LI and MA boundary coordinates, $\mathbf{c}_i$ is a vector of clinical covariates, and $y_i\in\{0,1\}$ is the available follow-up cardiovascular event label. Each ultrasound image was associated with a calibration factor $\kappa_i$ in mm/pixel, enabling the conversion of image-space measurements into physical units. For each participant, left and right common carotid artery scans were treated as image-level samples, whereas data splitting was performed at the participant level to avoid leakage between training, validation, and testing subsets.

The primary supervised task was defined as intima--media complex segmentation. A binary mask $\mathbf{M}_i\in\{0,1\}^{H\times W}$ was generated by filling the polygon enclosed by $\mathbf{B}^{LI}_i$ and $\mathbf{B}^{MA}_i$. The secondary supervised task was formulated as event-risk prediction using the follow-up event label. The joint learning objective was therefore defined as
\begin{equation}
    f_{\theta}: (\mathbf{I}_i,\mathbf{c}_i) \mapsto \left(\hat{\mathbf{M}}_i, \hat{p}_i, \hat{u}_i\right),
\end{equation}
where $\hat{\mathbf{M}}_i$ is the predicted intima--media probability map, $\hat{p}_i$ is the predicted cardiovascular risk probability, and $\hat{u}_i$ is the predictive uncertainty. In the full AtheroFlow-XNet formulation, $f_{\theta}$ is further extended to accept Doppler-derived flow descriptors and CFD-derived hemodynamic biomarker maps.

\subsection{Boundary-Derived Morphological Representation}
\label{subsec:morphology}

The LI and MA contours were used to construct a morphology-aware training target and to compute physical carotid wall descriptors. For a given image, the intima--media thickness profile was estimated by resampling the LI and MA boundaries along a common longitudinal coordinate. If $\mathbf{b}^{LI}(s)$ and $\mathbf{b}^{MA}(s)$ denote corresponding boundary points at normalized arc-length coordinate $s\in[0,1]$, the local image-space wall thickness was approximated as
\begin{equation}
    d_i(s)=\left\|\mathbf{b}^{MA}_i(s)-\mathbf{b}^{LI}_i(s)\right\|_2.
\end{equation}
The physical CIMT estimate was then computed as
\begin{equation}
    \mathrm{CIMT}_i = \kappa_i \cdot \frac{1}{S}\sum_{s=1}^{S} d_i(s),
\end{equation}
where $\kappa_i$ is the image-specific calibration factor. In addition to mean CIMT, spatial descriptors such as maximum thickness, thickness standard deviation, wall-area ratio, and boundary smoothness can be extracted from the segmentation output. These descriptors were treated as morphological surrogates of wall remodeling and were used to support clinical risk modeling.

\subsection{AtheroFlow-XNet Architecture}
\label{subsec:architecture}

The proposed AtheroFlow-XNet architecture was designed as a multi-branch model consisting of an ultrasound morphology encoder, a transformer-based contextual bottleneck, a decoder for LI--MA segmentation, a clinical fusion module, and an uncertainty-aware risk head. In the CUBS implementation, the morphology branch was trained directly from B-mode ultrasound images, while the clinical branch was trained from available demographic and vascular risk factors. The general full-model formulation can be written as
\begin{equation}
    \mathbf{z}_i = \Phi_{US}(\mathbf{I}_i) \oplus \Phi_{D}(\mathbf{d}_i) \oplus \Phi_{CFD}(\mathbf{h}_i) \oplus \Phi_{C}(\mathbf{c}_i),
\end{equation}
where $\Phi_{US}$ denotes the ultrasound image encoder, $\Phi_D$ denotes the Doppler-flow encoder, $\Phi_{CFD}$ denotes the hemodynamic biomarker encoder, $\Phi_C$ denotes the clinical covariate encoder, $\oplus$ indicates feature fusion, $\mathbf{d}_i$ represents Doppler-derived flow variables, and $\mathbf{h}_i$ represents CFD-derived biomarkers such as WSS, TAWSS, OSI, and RRT. Since $\mathbf{d}_i$ and $\mathbf{h}_i$ are not available in CUBS, the implemented training notebook uses
\begin{equation}
    \mathbf{z}_i = \Phi_{US}(\mathbf{I}_i) \oplus \Phi_C(\mathbf{c}_i),
\end{equation}
while preserving the modular interface required for Doppler and CFD extension.

The ultrasound encoder was implemented using residual convolutional blocks to extract speckle-robust local texture and boundary features. A transformer bottleneck was inserted at the deepest level of the encoder to capture long-range wall continuity and contextual dependencies along the carotid wall. The decoder followed an encoder--decoder topology with skip connections so that high-resolution boundary information could be preserved during segmentation. The segmentation head generated a pixel-wise probability map $\hat{\mathbf{M}}_i$, whereas the risk head generated $\hat{p}_i$ from the fused morphology--clinical latent representation. Dropout layers were retained during inference to enable Monte Carlo uncertainty estimation.

\subsection{Physics-Guided Hemodynamic Extension}
\label{subsec:physics_guided_extension}

Although CUBS does not include Doppler or CFD fields, the proposed framework was designed to be compatible with patient-specific hemodynamic modeling. When Doppler waveforms are available, the inlet flow profile can be represented as a time-dependent velocity or volumetric flow waveform $Q_{in}(t)$ derived from pulsed-wave Doppler. Patient-specific CFD can then be governed by the incompressible Navier--Stokes equations:
\begin{equation}
    \nabla \cdot \mathbf{u} = 0,
\end{equation}
\begin{equation}
    \rho\left(\frac{\partial \mathbf{u}}{\partial t}+\mathbf{u}\cdot\nabla\mathbf{u}\right) = -\nabla p + \nabla\cdot\left[\mu(\dot{\gamma})\left(\nabla\mathbf{u}+\nabla\mathbf{u}^{T}\right)\right],
\end{equation}
where $\mathbf{u}$ is blood velocity, $p$ is pressure, $\rho$ is blood density, and $\mu(\dot{\gamma})$ is the shear-rate-dependent apparent viscosity. A non-Newtonian Carreau--Yasuda model can be used when low-shear regions are expected to influence plaque-prone zones.

From the velocity gradient at the vessel wall, the instantaneous wall shear stress vector is given by
\begin{equation}
    \boldsymbol{\tau}_{w}(t)=\mu \left.\frac{\partial \mathbf{u}_{t}}{\partial n}\right|_{wall},
\end{equation}
where $\mathbf{u}_{t}$ is the tangential velocity component and $n$ is the wall-normal direction. The hemodynamic biomarkers are then computed as
\begin{equation}
    \mathrm{TAWSS}=\frac{1}{T}\int_{0}^{T}\left|\boldsymbol{\tau}_{w}(t)\right|dt,
\end{equation}
\begin{equation}
    \mathrm{OSI}=\frac{1}{2}\left(1-\frac{\left|\int_{0}^{T}\boldsymbol{\tau}_{w}(t)dt\right|}{\int_{0}^{T}\left|\boldsymbol{\tau}_{w}(t)\right|dt}\right),
\end{equation}
\begin{equation}
    \mathrm{RRT}=\frac{1}{(1-2\cdot\mathrm{OSI})\cdot\mathrm{TAWSS}}.
\end{equation}
These biomarkers were selected because previous patient-specific and CFD studies have shown that WSS-related quantities are associated with plaque burden, wall thickening, plaque composition, and ulceration risk [4,5,19,20]. Boundary condition selection was treated as a critical part of the pipeline because carotid CFD accuracy is strongly influenced by inlet and outlet assumptions [6,7].

To make the learning model physics-guided rather than purely data-driven, hemodynamic consistency terms can be added when CFD fields are available. The physics-guided loss is defined as
\begin{equation}
\begin{aligned}
    \mathcal{L}_{phys} = 
    &\ \lambda_{div}
    \left\|\nabla \cdot \hat{\mathbf{u}}\right\|_{2}^{2} \\
    &+ \lambda_{bc}
    \left\|\hat{Q}_{in}(t)-Q_{Doppler}(t)\right\|_{2}^{2} \\
    &+ \lambda_{wss}
    \left\|\hat{\boldsymbol{\tau}}_{w}
    -\boldsymbol{\tau}_{w}^{CFD}\right\|_{1}.
\end{aligned}
\end{equation}
where the three terms penalize violation of incompressibility, mismatch with Doppler-derived inlet flow, and inconsistency with CFD-derived WSS. This formulation is aligned with recent hybrid CFD--physics-informed learning principles [21], but is adapted to the carotid ultrasound domain.

\subsection{Uncertainty-Aware Risk Prediction}
\label{subsec:uncertainty}

Clinical translation requires not only a risk score but also an estimate of prediction confidence. Therefore, uncertainty was estimated using Monte Carlo dropout during inference. For each sample, $K$ stochastic forward passes were performed while dropout remained active:
\begin{equation}
    \left\{\hat{p}_{i}^{(k)},\hat{\mathbf{M}}_{i}^{(k)}\right\}_{k=1}^{K}=f_{\theta}^{(k)}(\mathbf{I}_i,\mathbf{c}_i).
\end{equation}
The final predictive probability and epistemic uncertainty were computed as
\begin{equation}
    \bar{p}_{i}=\frac{1}{K}\sum_{k=1}^{K}\hat{p}_{i}^{(k)},
\end{equation}
\begin{equation}
    u_i=\frac{1}{K}\sum_{k=1}^{K}\left(\hat{p}_{i}^{(k)}-\bar{p}_{i}\right)^2.
\end{equation}
For segmentation, the voxel-wise variance of $\hat{\mathbf{M}}_{i}^{(k)}$ was used to localize uncertain boundary regions. This uncertainty-aware mechanism was intended to improve interpretability by identifying cases in which risk prediction or wall-boundary estimation should be reviewed by clinicians.

\subsection{Training Objective}
\label{subsec:objective}

The total training objective was defined as a weighted multitask loss:
\begin{equation}
    \mathcal{L}_{total}=\lambda_{seg}\mathcal{L}_{seg}+\lambda_{risk}\mathcal{L}_{risk}+\lambda_{smooth}\mathcal{L}_{smooth}+\lambda_{phys}\mathcal{L}_{phys}.
\end{equation}
In the CUBS implementation, $\mathcal{L}_{phys}$ was disabled because Doppler and CFD fields were unavailable. The segmentation loss combined binary cross-entropy and Dice loss:
\begin{equation}
    \mathcal{L}_{seg}=\mathcal{L}_{BCE}(\hat{\mathbf{M}},\mathbf{M})+
    \left(1-\frac{2\sum \hat{\mathbf{M}}\mathbf{M}+\epsilon}{\sum \hat{\mathbf{M}}+\sum \mathbf{M}+\epsilon}\right).
\end{equation}
The event-risk loss was defined using binary cross-entropy:
\begin{equation}
    \mathcal{L}_{risk}=-y\log(\hat{p})-(1-y)\log(1-\hat{p}).
\end{equation}
Because follow-up labels are available only for a subset of the dataset, a label-availability mask was applied so that segmentation training could still use all annotated images. A smoothness regularizer was also used to encourage anatomically plausible wall-boundary probability maps:
\begin{equation}
    \mathcal{L}_{smooth}=\left\|\nabla_x \hat{\mathbf{M}}\right\|_1+\left\|\nabla_y \hat{\mathbf{M}}\right\|_1.
\end{equation}

\subsection{Training Protocol and Hyperparameters}
\label{subsec:hyperparameters}

All images were resized to $384\times384$ pixels for the baseline experiment. Patient-level group splitting was used to divide the dataset into training, validation, and testing subsets, preventing left--right image leakage from the same participant. Random horizontal flipping, intensity scaling, brightness shifting, and Gaussian noise injection were used as ultrasound-specific augmentations. The model was optimized using AdamW with a learning rate of $3\times10^{-4}$, weight decay of $1\times10^{-4}$, cosine learning-rate decay, and a warm-up period of five epochs. The baseline batch size was set to 8, the dropout rate was set to 0.20, and the maximum number of epochs was set to 60 in the provided training notebook. For publication-grade experiments, training should be repeated using multiple random seeds and a higher-resolution setting of $512\times512$ pixels when GPU memory permits.

The main hyperparameters used in the baseline implementation are summarized in Table~\ref{tab:hyperparameters}.

\begin{table}[htbp]
\centering
\caption{Baseline hyperparameters used for AtheroFlow-XNet-CUBS training.}
\label{tab:hyperparameters}
\begin{tabular}{ll}
\hline
\textbf{Hyperparameter} & \textbf{Value} \\
\hline
Input image size & $384\times384$ \\
Batch size & 8 \\
Optimizer & AdamW \\
Initial learning rate & $3\times10^{-4}$ \\
Minimum learning rate & $1\times10^{-6}$ \\
Weight decay & $1\times10^{-4}$ \\
Learning-rate schedule & Warm-up + cosine decay \\
Warm-up epochs & 5 \\
Training epochs & 60 baseline; 80--120 recommended \\
Dropout rate & 0.20 \\
Segmentation loss weight $\lambda_{seg}$ & 1.00 \\
Risk loss weight $\lambda_{risk}$ & 0.25 \\
Smoothness loss weight $\lambda_{smooth}$ & 0.03 \\
Monte Carlo dropout samples & 20 \\
Segmentation threshold & 0.50 \\
\hline
\end{tabular}
\end{table}

\subsection{Evaluation Metrics}
\label{subsec:evaluation}

Segmentation performance was evaluated using Dice similarity coefficient, intersection-over-union (IoU), boundary error, and CIMT error in millimeters. The Dice coefficient was computed as
\begin{equation}
    \mathrm{Dice}=\frac{2|\hat{\mathbf{M}}\cap\mathbf{M}|}{|\hat{\mathbf{M}}|+|\mathbf{M}|}.
\end{equation}
For the risk-prediction task, area under the receiver operating characteristic curve (AUC), accuracy, F1-score, sensitivity, specificity, and calibration error were recommended. Uncertainty quality was evaluated by examining whether high-uncertainty cases corresponded to lower segmentation accuracy, ambiguous boundaries, or inconsistent risk predictions. Explainable risk maps were produced from predicted segmentation probability, uncertainty maps, and activation-based saliency over the carotid wall region.

\subsection{Implementation Details}
\label{subsec:implementation}

The training notebook accompanying this study implements the full CUBS-compatible pipeline in PyTorch. The notebook automatically extracts the CUBS zip file, constructs LI--MA segmentation masks from manual contours, reads calibration factors, builds patient-level train--validation--test splits, trains the AtheroFlow-XNet-CUBS model, evaluates segmentation and risk-prediction performance, and performs Monte Carlo dropout inference for uncertainty estimation. This implementation is intended as a reproducible baseline that can be extended with Doppler-derived inlet waveforms and CFD-derived WSS, TAWSS, OSI, and RRT maps in future experiments.

\section{Results and Analysis}

\subsection{Dataset Partitioning and Experimental Setting}

The proposed \textit{AtheroFlow-XNet} model was evaluated using the Carotid Ultrasound Boundary Study (CUBS) dataset, an open multicenter carotid ultrasound dataset released by Meiburger et al. [1]. The dataset contains B-mode carotid ultrasound images with manually annotated lumen--intima (LI) and media--adventitia (MA) boundaries, calibration factors, and associated clinical information. In this study, the LI and MA boundary coordinates were converted into pixel-level intima--media masks and used as the primary supervision target for ultrasound-based carotid wall segmentation. The clinical variables were incorporated as auxiliary patient-level descriptors for the risk-prediction branch.

To avoid patient-level information leakage, the dataset was divided at the patient level rather than the image level. The final split consisted of 1,522 training images from 761 patients, 326 validation images from 163 patients, and 328 testing images from 164 patients. Each B-mode ultrasound image and its corresponding binary LI--MA mask were resized to $384 \times 384$ pixels. During training, each batch contained ultrasound images, segmentation masks, normalized clinical variables, binary risk labels, and label masks. The resulting tensor dimensions were verified as follows: image tensors of size $8 \times 1 \times 384 \times 384$, mask tensors of size $8 \times 1 \times 384 \times 384$, clinical feature tensors of size $8 \times 5$, and label tensors of size $8 \times 1$.

\begin{figure*}[h]
    \centering
    \includegraphics[width=0.7\linewidth]{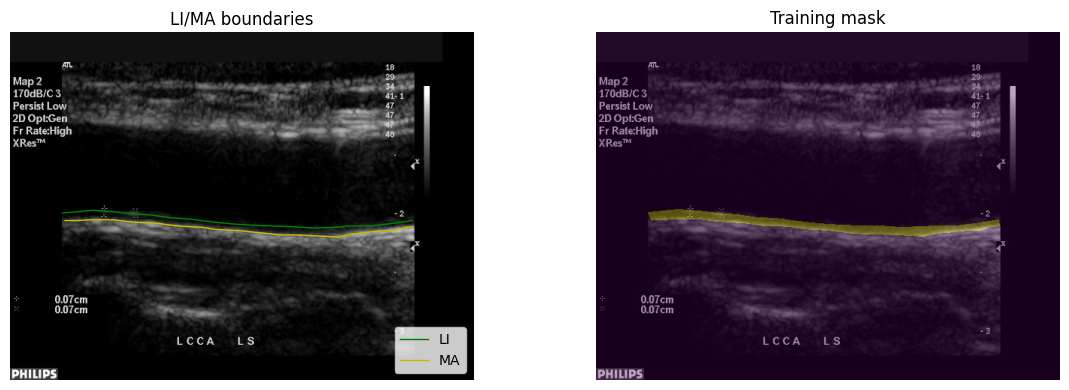}
    \caption{
    Boundary-derived mask generation from manual LI and MA annotations. 
    The lumen--intima (LI) and media--adventitia (MA) boundaries were extracted from the CUBS manual annotations, and the region between both contours was filled to generate the dense intima--media training mask. 
    This conversion reformulated the original boundary annotation task as a pixel-level carotid wall segmentation problem.
    }
    \label{fig:mask_generation}
\end{figure*}

The conversion from boundary annotations to dense training masks is illustrated in Fig.~\ref{fig:mask_generation}. The LI and MA contours were used to define the intima--media region, and the area enclosed by both boundaries was filled to produce a binary supervision mask. This preprocessing step reformulated the original CUBS boundary annotation problem into a pixel-level carotid wall segmentation task.

\subsection{Segmentation Performance}

The segmentation branch of \textit{AtheroFlow-XNet} was evaluated by comparing the predicted intima--media masks with the manually derived LI--MA masks. The model achieved a segmentation loss of 0.2359 and a Dice similarity coefficient of 0.7930 on the evaluation set. This result indicates that the proposed model was able to learn a consistent representation of the carotid intima--media region from B-mode ultrasound images, despite the low contrast, speckle noise, shadowing artifacts, and inter-device variability commonly observed in ultrasound imaging.

\begin{figure}[h]
    \centering
    \includegraphics[width=0.5\textwidth]{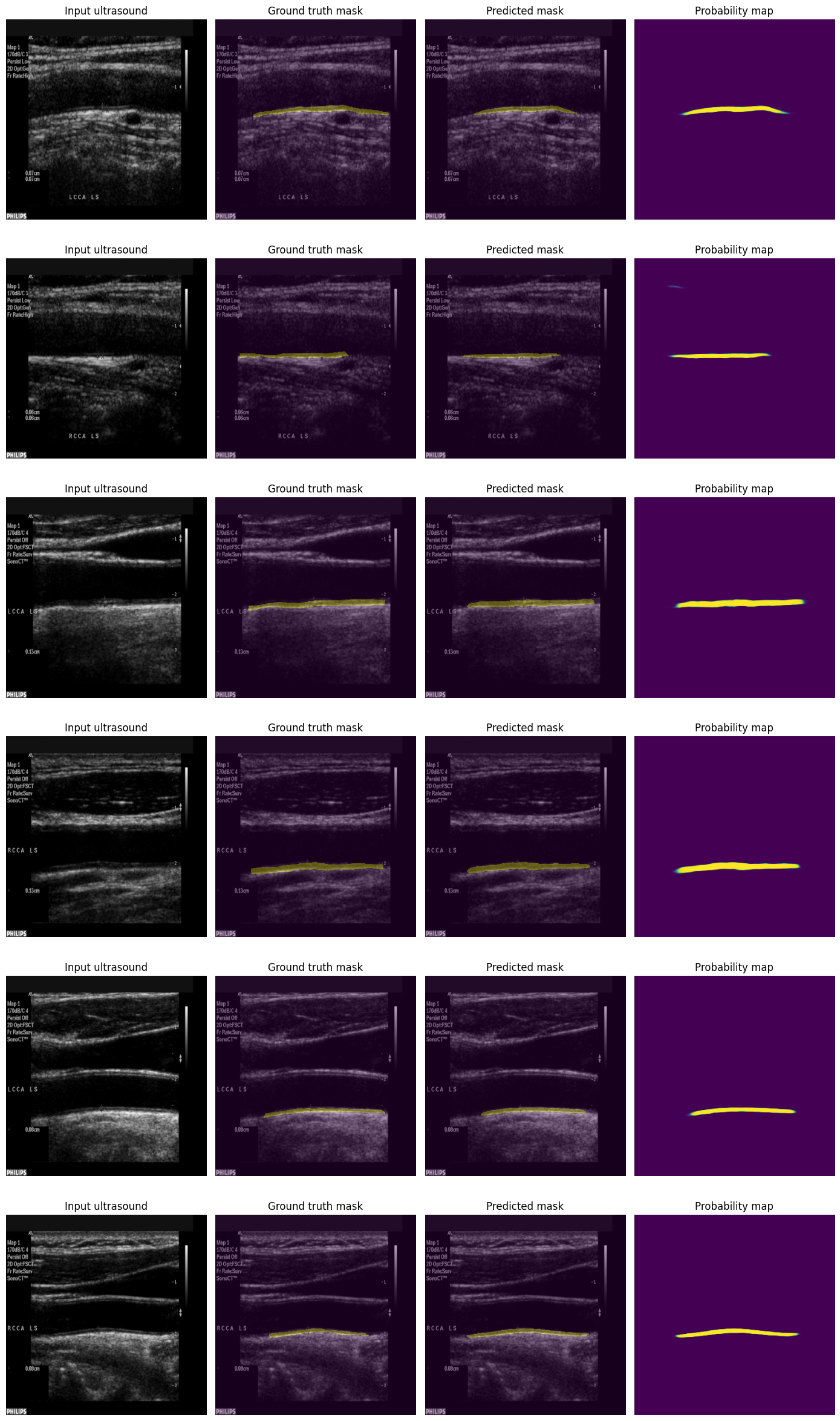}
    \caption{
    Qualitative segmentation results of the proposed \textit{AtheroFlow-XNet} model on CUBS carotid ultrasound images. 
    Each row shows the input B-mode ultrasound image, the manually derived LI--MA ground-truth mask, the predicted mask, and the corresponding probability map. 
    The predicted intima--media regions were generally aligned with the manual annotations, indicating that the model captured the thin carotid wall structure under ultrasound noise and image-quality variations.
    }
    \label{fig:segmentation_results}
\end{figure}

Qualitative segmentation results are shown in Fig.~\ref{fig:segmentation_results}. The predicted masks closely followed the manually annotated LI--MA regions across multiple carotid ultrasound views. In most cases, the predicted intima--media band was spatially aligned with the ground-truth mask, particularly along continuous arterial wall segments. The corresponding probability maps showed high-confidence activation concentrated along the expected intima--media region, while background tissues, lumen regions, and deeper soft-tissue structures were largely suppressed. These observations suggest that the model learned anatomically meaningful wall-related features rather than relying only on global image intensity patterns.

\subsection{Uncertainty-Aware Prediction Analysis}

To evaluate the confidence behavior of the model, Monte Carlo dropout was applied during inference. Multiple stochastic forward passes were performed for each test image, and the mean prediction and predictive variance were computed. The mean risk predictions for representative samples were 0.0242, 0.0535, 0.0454, 0.2211, and 0.0202, while the corresponding predictive uncertainties were 0.00011, 0.00064, 0.00081, 0.00703, and 0.00015, respectively. These results indicate that most predictions were associated with low uncertainty, whereas relatively higher uncertainty was observed in samples with higher predicted risk probability.

\begin{figure*}[h]
    \centering
    \includegraphics[width=\textwidth]{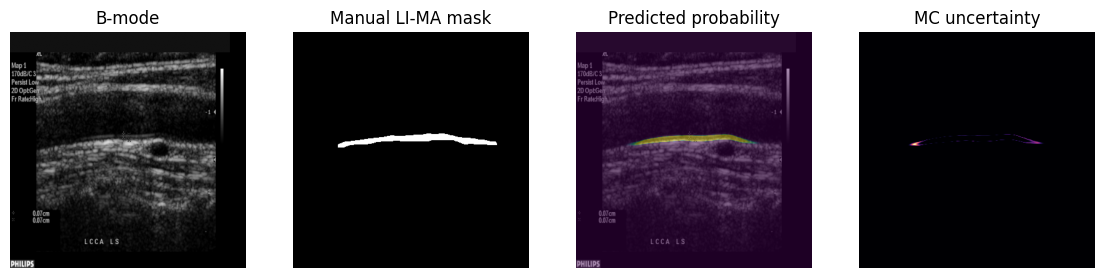}
    \caption{
    Representative uncertainty-aware prediction generated by Monte Carlo dropout inference. 
    The input B-mode ultrasound image, manual LI--MA mask, predicted probability map, and predictive uncertainty map are shown. 
    The uncertainty response was mainly localized near ambiguous wall-boundary regions, suggesting that the model confidence was spatially related to boundary clarity.
    }
    \label{fig:uncertainty_single_case}
\end{figure*}

A representative uncertainty-aware prediction is shown in Fig.~\ref{fig:uncertainty_single_case}. The B-mode ultrasound image, manual LI--MA mask, predicted probability map, and Monte Carlo uncertainty map are presented for one sample. The predicted probability was concentrated along the intima--media region, while uncertainty was mainly localized near boundary regions where the wall interface was less distinct.

\begin{figure}[h]
    \centering
    \includegraphics[width=0.5\textwidth]{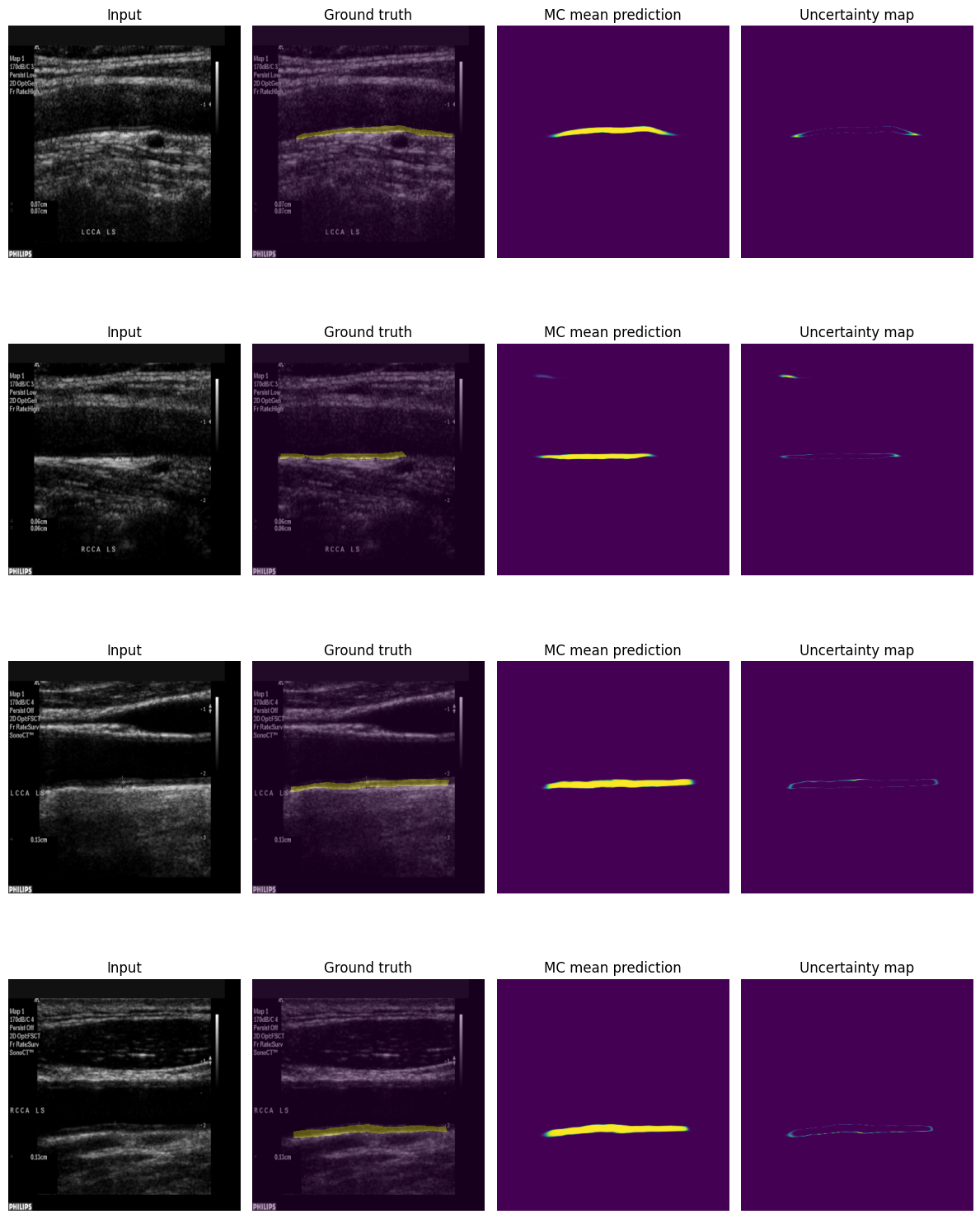}
    \caption{
    Monte Carlo dropout-based uncertainty estimation for carotid intima--media segmentation across multiple CUBS test samples. 
    For each sample, the input ultrasound image, ground-truth mask, Monte Carlo mean prediction, and predictive uncertainty map are shown. 
    The predictive uncertainty was mainly concentrated near boundary regions and locally ambiguous wall interfaces, indicating that uncertainty-aware inference can highlight anatomically difficult regions for expert inspection.
    }
    \label{fig:uncertainty_results}
\end{figure}

The uncertainty behavior across multiple test samples is further visualized in Fig.~\ref{fig:uncertainty_results}. The uncertainty maps demonstrate that predictive uncertainty was not uniformly distributed across the entire ultrasound image. Instead, uncertainty was concentrated near the predicted intima--media boundary and at local regions where the wall interface was less distinct. This behavior is desirable for clinical image analysis because ambiguous boundary regions, acoustic artifacts, and weak wall contrast are expected to produce greater predictive uncertainty. Therefore, the uncertainty branch provided additional information beyond the deterministic mask output and may be useful for identifying cases that require manual review or expert confirmation.

\subsection{Risk Prediction Performance}

In addition to segmentation, the proposed framework included a risk-prediction branch that integrated image-derived representation and clinical features. The model achieved an area under the receiver operating characteristic curve (AUC) of 0.6910. Although this value indicates moderate discriminative performance, it should be interpreted carefully because the CUBS dataset was originally designed for CIMT boundary analysis and clinical impact assessment rather than for direct plaque vulnerability or hemodynamic risk stratification. Therefore, the available labels are more suitable for demonstrating feasibility of ultrasound-based morphology learning than for definitive high-risk plaque classification.

The moderate AUC suggests that B-mode morphology and limited clinical variables contain useful but incomplete information for patient-level vascular risk estimation. This finding is consistent with the central motivation of \textit{AtheroFlow-XNet}: morphology alone may not be sufficient for robust atherosclerosis risk stratification. Additional information, including Doppler-derived flow profiles, CFD-derived wall shear stress biomarkers, plaque composition, stenosis severity, and longitudinal outcomes, would likely improve the discriminative capacity of the proposed framework. Thus, the current results should be regarded as a CUBS-compatible baseline demonstrating the feasibility of the ultrasound morphology and uncertainty-aware learning components.

\begin{table}[t]
\centering
\caption{Quantitative performance of the proposed \textit{AtheroFlow-XNet} model on the CUBS-based evaluation setting.}
\label{tab:quantitative_results}
\begin{tabular}{lc}
\hline
\textbf{Metric} & \textbf{Value} \\
\hline
Segmentation loss & 0.2359 \\
Dice coefficient & 0.7930 \\
Risk prediction AUC & 0.6910 \\
Training images / patients & 1522 / 761 \\
Validation images / patients & 326 / 163 \\
Testing images / patients & 328 / 164 \\
Input image size & $384 \times 384$ \\
Clinical feature dimension & 5 \\
\hline
\end{tabular}
\end{table}

\subsection{Clinical and Methodological Interpretation}

The obtained results demonstrate that \textit{AtheroFlow-XNet} can provide three clinically relevant outputs from carotid ultrasound data: intima--media segmentation, risk probability estimation, and uncertainty-aware prediction. The segmentation result is particularly important because accurate localization of the intima--media region is a prerequisite for morphology-based vascular assessment. A Dice coefficient of 0.7930 indicates that the model achieved a meaningful overlap with manual annotations, while the qualitative examples show that the predicted masks were anatomically plausible across different image appearances.

The uncertainty results further strengthen the interpretability of the framework. In clinical applications, deterministic predictions alone may be insufficient because ultrasound images are affected by operator variability, acoustic shadowing, poor contrast, and anatomical heterogeneity. By estimating predictive uncertainty, the model can provide an additional confidence measure that may support human-in-the-loop decision-making. Cases with high uncertainty may be flagged for manual verification, while low-uncertainty cases may be considered more reliable for automated analysis.

Nevertheless, the current CUBS-based experiment represents only the first stage of the full \textit{AtheroFlow-XNet} framework. Since the CUBS dataset does not directly provide Doppler waveform data, three-dimensional CFD fields, WSS, TAWSS, OSI, or RRT maps, the hemodynamic learning components were not fully evaluated in this experiment. Therefore, the current implementation should be interpreted as an ultrasound morphology-driven baseline. In future work, Doppler-informed boundary conditions and CFD-derived wall shear biomarkers should be incorporated to complete the proposed physics-guided multiscale hemodynamic learning pipeline.

\subsection{Limitations}

Several limitations should be noted. First, the CUBS dataset was primarily designed for computerized CIMT measurement and clinical impact analysis rather than for direct atherosclerotic plaque vulnerability prediction. Consequently, the available labels may not fully represent patient-specific hemodynamic risk or future cerebrovascular events. Second, only B-mode ultrasound images, LI--MA boundary-derived masks, and limited clinical features were used in the current experiment. Doppler waveforms and CFD-derived biomarkers were not available in the dataset and therefore could not be directly integrated into the present evaluation. Third, the risk-prediction AUC of 0.6910 indicates moderate performance, suggesting that morphology-only learning is insufficient for comprehensive risk stratification. Fourth, external validation on independent ultrasound datasets and prospective clinical cohorts is still required to assess generalizability.

Despite these limitations, the results support the feasibility of the proposed framework as a foundation for ultrasound-based carotid wall analysis. The segmentation branch provided anatomically consistent LI--MA localization, and the uncertainty-aware inference mechanism generated meaningful confidence information. These components can be extended in future studies by incorporating Doppler-derived flow conditions, patient-specific CFD simulations, and hemodynamic biomarkers such as WSS, TAWSS, OSI, and RRT.

\subsection{Summary of Findings}

In summary, the proposed \textit{AtheroFlow-XNet} model was successfully trained and evaluated on the CUBS dataset. The model achieved a Dice coefficient of 0.7930 for LI--MA mask segmentation and an AUC of 0.6910 for risk prediction. Qualitative analysis showed that the predicted masks were well aligned with manual LI--MA annotations, while Monte Carlo dropout-based uncertainty maps highlighted ambiguous boundary regions. These findings suggest that ultrasound-derived carotid morphology can be effectively learned by the proposed model and that uncertainty-aware prediction can enhance interpretability. However, the moderate risk-prediction performance also indicates that full patient-specific atherosclerosis risk stratification will likely require additional Doppler and CFD-derived hemodynamic information.

\section{Conclusion}

In this study, an uncertainty-aware physics-guided multiscale learning framework, termed \textit{AtheroFlow-XNet}, was proposed for patient-specific carotid atherosclerosis risk stratification. The framework was designed to integrate ultrasound-derived vascular morphology, clinical descriptors, uncertainty-aware prediction, and an extensible hemodynamic learning formulation that can incorporate Doppler-informed flow profiles and CFD-derived wall shear biomarkers when such data are available. Using the CUBS dataset, the implemented model was evaluated as a morphology-driven baseline for carotid intima--media segmentation and risk prediction.

The experimental results demonstrated that the proposed model was able to learn anatomically meaningful carotid wall representations from B-mode ultrasound images. The segmentation branch achieved a Dice coefficient of 0.7930 and a segmentation loss of 0.2359, indicating that the predicted LI--MA masks were generally well aligned with manual annotations. Qualitative visualization further showed that the predicted probability maps were concentrated along the intima--media region, suggesting that the model was able to identify thin carotid wall structures despite ultrasound speckle noise, low boundary contrast, and image-quality variability. In addition, Monte Carlo dropout-based uncertainty estimation showed that predictive uncertainty was mainly localized near ambiguous wall-boundary regions, supporting the potential usefulness of uncertainty-aware inference for human-in-the-loop clinical review.

For risk prediction, the model achieved an AUC of 0.6910. This moderate discriminative performance suggests that B-mode morphology and limited clinical features provide useful but incomplete information for patient-specific vascular risk estimation. Since the CUBS dataset was primarily designed for computerized CIMT measurement and clinical impact analysis rather than direct plaque vulnerability or hemodynamic risk stratification, the present experiment should be interpreted as a feasibility baseline rather than a definitive clinical risk model. The result also supports the central motivation of \textit{AtheroFlow-XNet}: morphology-only learning may be insufficient for comprehensive atherosclerosis risk assessment, and additional physiological information is required.

Several limitations remain. First, Doppler waveform data, patient-specific three-dimensional vascular geometries, and CFD-derived hemodynamic biomarkers such as WSS, TAWSS, OSI, and RRT were not directly available in the CUBS dataset. Therefore, the full hemodynamic branch of \textit{AtheroFlow-XNet} could not be experimentally validated in the present study. Second, the available risk labels may not fully represent plaque vulnerability, stenosis progression, or future cerebrovascular events. Third, external validation using independent multicenter datasets and prospective clinical cohorts is required before clinical translation can be considered.

Future work will focus on extending the current framework by incorporating Doppler-derived inlet flow profiles, patient-specific vascular reconstruction, and CFD-based wall shear biomarkers. The integration of these hemodynamic descriptors is expected to improve the physiological interpretability and discriminative performance of the model. Moreover, external validation, calibration analysis, and clinician-centered explainability evaluation should be performed to assess the robustness and clinical utility of the proposed framework. Overall, the present study establishes a reproducible ultrasound-based foundation for \textit{AtheroFlow-XNet} and demonstrates its potential as a step toward personalized, uncertainty-aware, and mechanism-informed carotid atherosclerosis risk assessment.

\section*{Acknowledgment}

The author would like to acknowledge the contributors of the Carotid Ultrasound Boundary Study (CUBS) dataset for providing an open multicenter carotid ultrasound resource that enabled the development and evaluation of the proposed framework. The author also gratefully acknowledges the School of Information, Computer, and Communication Technology, Sirindhorn International Institute of Technology, Thammasat University, for providing academic support and a research environment for this work. Appreciation is extended to the open-source scientific computing community, including the developers of PyTorch, OpenCV, NumPy, pandas, scikit-learn, and Matplotlib, whose tools supported the implementation, training, visualization, and analysis of the proposed \textit{AtheroFlow-XNet} pipeline.

\appendices

\section{Reproducibility and Experimental Artifacts}

To support reproducibility, all experiments were conducted using a fixed random seed of 42. The best-performing model was saved as a PyTorch checkpoint file named \texttt{atheroflow\_xnet\_cubs\_best.pth}. The checkpoint contained three main components: the trained model parameters, the training configuration, and the training history. During inference, the model state dictionary was extracted from the checkpoint key \texttt{model}. Successful checkpoint loading was verified by confirming that no missing or unexpected parameter keys were reported.

The model was trained using the patient-level train--validation--test split described in Section~\ref{sec:methodology}. The final split consisted of 1,522 training images from 761 patients, 326 validation images from 163 patients, and 328 testing images from 164 patients. All images were resized to $384\times384$ pixels, and the clinical feature vector contained five variables: age, sex, hypertension, diabetes, and body mass index. These details were retained in the saved configuration to ensure that the same preprocessing and inference settings could be reproduced.

\section{Output Files and Visualization Protocol}

The experimental pipeline generated quantitative metric files and qualitative visualization outputs for model inspection. The main output files included segmentation metrics, risk-prediction metrics, prediction tables, segmentation examples, Monte Carlo dropout uncertainty maps, and representative single-case uncertainty visualizations. The qualitative figures used in the results section were generated from held-out test samples only.

For visual inspection, each segmentation example was displayed using four components: the input B-mode ultrasound image, the manually derived LI--MA ground-truth mask, the predicted binary mask, and the predicted probability map. For uncertainty analysis, Monte Carlo dropout inference was performed using 20 stochastic forward passes. The resulting visualization included the input image, ground-truth mask, Monte Carlo mean prediction, and predictive uncertainty map. These outputs were used only for interpretation and were not used to tune the final test-set metrics.

\section{Code Availability Statement}

The implementation was developed in Python using PyTorch, OpenCV, NumPy, pandas, scikit-learn, and Matplotlib. The notebook pipeline was structured to perform dataset loading, LI--MA mask construction, patient-level splitting, model training, checkpoint saving, evaluation, and uncertainty visualization. The code can be extended to include Doppler-derived inlet waveforms and CFD-derived WSS, TAWSS, OSI, and RRT maps when such data are available.

\end{document}